\documentclass{ifacconf}

\usepackage{graphicx}      
\usepackage{natbib}        
\usepackage{amssymb}
\usepackage{amstext}
\usepackage{amsmath,bm}
\usepackage{mathtools}
\usepackage{xspace}
\usepackage{graphicx}
\usepackage{algorithm}
\usepackage{algpseudocode}
\usepackage{tikz}
\usepackage{url}
\usepackage{pgf}
\usepackage{tikz}
\usepackage{graphicx}
\usepackage{subcaption}

\usetikzlibrary{positioning, arrows.meta, fit}

\usepackage{xcolor,soul}

\theoremstyle{definition}

\usetikzlibrary{shapes,shadows,arrows,positioning,angles,quotes}
\tikzset{My Arrow Style/.style={single arrow, fill=black!15, anchor=base, align=center,text width=2.3cm}}
\tikzstyle{arrow} = [thick,->,>=stealth]
\usetikzlibrary{calc,trees,positioning,arrows,fit,shapes}

\tikzstyle{startstop} = [rectangle, rounded corners, minimum width=1.5cm, minimum height=0.5cm,text centered, draw=black, fill=red!30]
\tikzstyle{io} = [trapezium, trapezium left angle=70, trapezium right angle=110, minimum width=1cm, minimum height=0.5cm, text centered, draw=black, fill=blue!30]

\tikzstyle{process} = [rectangle, minimum width=3cm, minimum height=0.5cm, text centered, draw=black, fill=orange!30]
\tikzstyle{decision} = [diamond, minimum width=1cm, minimum height=0.5cm, text centered, draw=black, fill=green!30]

\tikzstyle{arrow} = [thick,->,>=stealth]

\begin{document}
\begin{frontmatter}

\title{Efficient Milling Quality Prediction with Explainable Machine Learning\thanksref{footnoteinfo}} 

\thanks[footnoteinfo]{This work is funded by the European Union under grant agreement number 101091783 (MARS Project) and as part of the Horizon Europe HORIZON-CL4-2022-TWIN-TRANSITION-01-03.}

\author[First]{Dennis Gross} 
\author[First]{Helge Spieker} 
\author[First]{Arnaud Gotlieb}
\author[Second]{Ricardo Knoblauch}
\author[Second]{Mohamed Elmansori}

\address[First]{Simula Research Laboratory, Oslo, Norway\\ (e-mail: dennis@simula.no).}
\address[Second]{Ecole Nationale Supérieure d'Arts et Métiers, Aix-en-Provence, France.}

\begin{abstract}                
This paper presents an explainable machine learning (ML) approach for predicting surface roughness in milling. Utilizing a dataset from milling aluminum alloy 2017A, the study employs random forest regression models and feature importance techniques. The key contributions include developing ML models that accurately predict various roughness values and identifying redundant sensors, particularly those for measuring normal cutting force. 
Our experiments show that removing certain sensors can reduce costs without sacrificing predictive accuracy, highlighting the potential of explainable machine learning to improve cost-effectiveness in~machining.
\end{abstract}

\begin{keyword}
Machine Learning, Milling, Sustainable Production, Explainable ML
\end{keyword}

\end{frontmatter}

\section{Introduction}
\emph{Machine Learning (ML)} significantly impacts the manufacturing industry~\citep{DBLP:journals/access/JyeniskhanKKAS23,DBLP:journals/ijcim/Jiang23}.
Applying ML to manufacturing offers improved efficiency~\citep{panzer2022deep}, predictive maintenance~\citep{DBLP:journals/eswa/WangLLLZ23}, and better control over manufacturing quality~\citep{DBLP:journals/candie/KimCKL23}.

In general, ML uses algorithms to interpret data and make predictions.
It involves creating models that learn from a training data set and can be used to make predictions over unseen data. 
Before deployment, the trained ML model is tested on a test data set~\citep{DBLP:journals/ijcim/Jiang23}.

However, integrating trained ML models into industrial applications comes with challenges~\citep{DBLP:journals/sensors/SampedroRKL22,DBLP:journals/corr/abs-2310-09991}.
One of these challenges is the "black box" nature of many ML models, making it difficult for human experts to trust the prediction results of trained ML models~\citep{DBLP:journals/cogsr/KwonKC23}.
This is essential because, in manufacturing, the addition of sensors to a machine tool for generating data for the ML model can be costly, difficult to be implemented (necessity of trained staff), may interfere in the working area inside the machine, and even might modify machine's behaviour (e.g. reducing machine rigidity with installation of a dynamometer).~\citep{dornfeld2008sensors,DBLP:journals/arc/HawkridgeMMTPRT21}.
The inclusion of a specific sensor value in the data collection phase for a prototype system does not necessarily imply its significance for the ML model's predictive capabilities; hence, it might be reasonable to consider omitting this sensor feature in the real system if subsequent analysis shows it does not contribute meaningfully to the model's performance.
\emph{Explainable ML} encompasses methods that render the outputs of ML models comprehensible to humans, allowing for the analysis of how various features contribute to the model's predictions~\citep{DBLP:journals/cbm/RasheedQGARQ22,DBLP:journals/ai/TiddiS22,DBLP:journals/access/TheisslerSSG22}.

In this paper, we showcase with explainable ML methods that it is possible to train explainable ML models and to identify and remove already mounted non-significant sensors for high-quality roughness predicting ML models in the context of a milling system.
The dataset for this paper was generated at MSMP - ENSAM, encompassing a series of surface milling operations on aluminium alloy 2017A.
These operations employed a 20 mm diameter milling cutter, specifically the R217.69-1020.RE-12-2AN model equipped with two XOEX120408FR-E06 H15 carbide inserts from SECO.
The process utilized a synthetic emulsion comprising water and 5\% Ecocool CS+ cutting~fluid.

The rest of the paper is organized as follows: Section~\ref{sec:related} covers existing works on using AI for manufacturing/machining problems; Section~\ref{sec:metho} presents our explainable ML methodology; Section~\ref{sec:case} provides a comprehensive guide on harnessing explainable machine learning techniques for the milling data set.
In Section~\ref{sec:discu} and Section~\ref{sec:conclu}, we delve into the advantages and limitations of employing explainability methods in machining.
These sections also offer concluding remarks and insights regarding the future prospects of this research.

\section{Related Work}\label{sec:related}
The usage of ML in manufacturing/machining tasks has been recognized as an interesting lead for at least a decade \citep{Kummar17}. For instance, ML has been used initially to optimize turning processes \citep{mokhtari2014optimization}, predicting stability conditions in milling \citep{postel2020ensemble}, estimating the quality of bores \citep{schorr20}, or classifying defects using ML-driven surface quality control \citep{Chouhad21}.

However, it is only recently that Explainable AI (XAI) methods have been identified as an interesting approach for manufacturing processes \citep{Yoo21,DBLP:journals/mansci/SenonerNF22}.
The ongoing European XMANAI project \citep{Lampathaki21} aims to evaluate the capabilities of XAI in different sectors of manufacturing through the development of several use cases.
In particular, fault diagnosis seems to be an area where XAI can be successfully applied \citep{Brusa23}. Also, there exists work that focuses on feature selection on the dataset without taking the ML model directly into account~\citep{bins2001feature,oreski2017effects,venkatesh2019review}.
Using a simple milling dataset, we initially showed in \citep{DBLP:conf/icaart2024/GrossS0023} that ML models such as decision tree regression, gradient boosting regression, and random forest lead to interesting performances for accurate roughness value prediction. In \citep{DBLP:conf/icaart2024/GrossS0023}, we also compared different explainability methods and observed that unfortunately, different methods lead to different explanations. This result calls for more experiments and results on the potential interest and benefice of using XAI methods in machining. 
Our paper specifically focuses on developing ML models for predicting roughness in milling and identifying redundant sensors.
In contrast, \citep{DBLP:conf/icaart2024/GrossS0023} broadly aims to enhance the performance of ML models in forecasting milling quality through explainable machine learning methods.

\section{Methodolodgy}\label{sec:metho}
This study aims to develop an explainable ML model for predicting milling surface roughness.
Our approach focuses on using random forest regression (see Section~\ref{sec:rfm}) and feature importance methods (see Section~\ref{sec:ex}) to not only achieve accurate and explainable predictions but also identify and eliminate non-essential sensors, thus enhancing cost-effectiveness.

\subsection{Random Forest Regression Models}\label{sec:rfm}
Random Forest Regression models also employ an ensemble learning strategy, building multiple decision trees during the training phase and aggregating them for predictions. The final prediction \( \hat{y} \) for an input \( x \) is the average prediction across all trees in the ensemble:

\[
\hat{y}(x) = \frac{1}{T} \sum_{t=1}^{T} y_t(x)
\]
where \( T \) is the total number of trees and \( y_t(x) \) is the prediction of the \( t \)-th tree. This aggregation helps enhance the model's generalization capabilities and mitigates the risk of overfitting~\citep{prasad2006newer}. {\it Overfitting} is an undesirable ML behaviour occurring when the model fits too precisely the dataset used for its training and reveals itself incapable of generalizing properly to unseen data.

\subsection{Explainable ML Methods}\label{sec:ex}
A critical aspect of our approach is employing feature permutation importance as a significant explainability method. This technique operates by evaluating the importance of different features in the model. The general procedure involves the random permutation of a single feature, keeping others constant, and monitoring the change in the model's performance, often measured through metrics like mean squared error~\citep{huang2016permutation,DBLP:conf/icaart2024/GrossS0023}.

Mathematically, the feature importance \( I_i \) of a feature \( i \) can be defined as the difference in the model's performance before and after the permutation of the feature and can be formulated as:
\[ I_i = P_{original} - P_{permuted(i)} \]
where \( P_{original} \) is the model's performance with the original data and \( P_{permuted(i)} \) is the performance with the \(i\)-th feature permuted.

By iterating this process across all features and comparing the changes in performance, we can rank the features by their importance, offering deeper insights into the model's decision-making process and enabling the identification of areas for optimization and refinement.

In Scikit-learn's Random Forest Regression~\citep{scikit-learn}, the importance of a feature is the Gini importance~\citep{menze2009comparison}.

\section{Case Study}\label{sec:case} 
In this case study, we apply our explainable ML to a dataset generated at MSMP - ENSAM.
The dataset consists of a series of surface milling operations that were performed on aluminium 2017A using a 20 mm diameter milling cutter R217.69-1020.RE-12-2AN with two carbide inserts XOEX120408FR-E06 H15 from SECO, and a synthetic emulsion of water and 5\% of Ecocool CS+ (~5\%) cutting~fluid. 

\emph{Objective.}
The aim is to develop a predictive model for each quality metric associated with roughness amplitude parameters.
This necessitates not only the training of accurate models but also an elucidation of the predictive rationales behind their outputs.
Concurrently, there is a need to identify and eliminate superfluous features from the models.
This is a strategic step to minimize both installation and maintenance expenses related to redundant sensors, thereby optimizing resource allocation and reducing overall costs.

\emph{Dataset.} In total, 200 experiments have been carried out varying the following process parameters: depth of cut, cutting speed, feed rate, and cutting mode (down and up milling).
For each one of these experiments with different control parameters, cutting forces $F_z$ (normal force) and $F_a$ (active force), and surface profiles are measured on-machine using a Kistler 3-axis dynamometer 9257A and a STIL CL1-MG210 chormatic confocal sensor (non-contact) respectively.
In the feature engineering step, the following surface roughness parameters are calculated using MountainsMap software.
Some of the quality parameters are: 
\begin{itemize}
    \item Ra (Average Roughness): Average value of the absolute distances from the mean line to the roughness profile within the evaluation length. 
    \item Rz (Average Maximum Height): Average value of the five highest peaks and the five deepest valleys within the evaluation length. 
    \item Rt (Total Roughness): Vertical distance between the highest peak and the deepest valley within the evaluation length. 
    \item Rq (Root Mean Square Roughness): Square root of the average of the squared distances from the mean line to the roughness profile within the evaluation length.
    \item RSm (Mean Summit Height): Average height of the five highest peaks within the evaluation length. 
    \item RSk (Skewness): Measure of the asymmetry of the roughness profile around the mean line. 
    \item Rku (Kurtosis): Measure the peakedness or flatness of the roughness profile. 
    \item Rmr (Material Ratio): Ratio of the actual roughness profile area to the area within the evaluation length. 
    \item Rpk (Peak Height): Height of the highest peak within the evaluation length. 
    \item Rvk (Valley Depth): Depth of the deepest valley within the evaluation length. 
    \item Rdq: It is a hybrid parameter (height and length). It is the root mean square slope of the assessed profile, defined on the sampling length. Rdq is the first approach to surface complexity. A low value is found on smooth surfaces while higher values can be found on rough surfaces having microroughness.
\end{itemize}

\emph{Data Preprocessing.}
Since we are dealing with variable time series lengths, we calculate the box plot values for each time series in the time and frequency domain.
Additionally, metadata within the dataset comprises experiment parameters with various focuses.

\begin{figure}[t]
    \centering
    \includegraphics[width=0.4\textwidth]{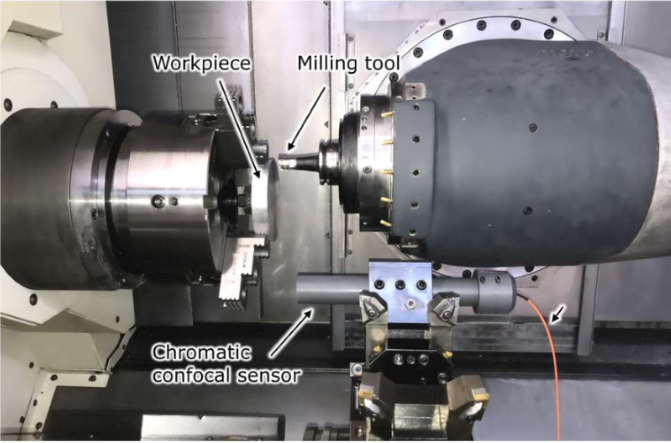}
    \caption{Milling machine that produces workpieces.}
    \label{fig:machine1}
\end{figure}

\begin{figure}[]
\centering
    \scalebox{0.9}{
    \begin{tikzpicture}[>=Stealth, node distance=0.25cm]
        \node (input1) {$F_A$ Box Plot};
        \node [below=0.1cm of input1] (input2) {$F_Z$ Box Plot};
        \node [below=0.1cm of input2] (input3) {f [mm/rot of the tool]};
        \node [below=0.1cm of input3] (input4) {n [rpm]};
        \node [below=0.1cm of input4] (input5) {vc [m/min]};
        \node [below=0.1cm of input5] (input6) {ap [mm]};
        
        \node [right=0.7cm of input3, draw, rounded corners, inner sep=20pt, minimum height=4cm, yshift=-0.3cm] (model) {Model};

        \node [right=0.5cm of model] (output) {Quality};
    
        \draw[->] (input1) -- (model);
        \draw[->] (input2) -- (model);
        \draw[->] (input3) -- (model);
        \draw[->] (input4) -- (model);
        \draw[->] (input5) -- (model);
        \draw[->] (input6) -- (model);
        
        \draw[->] (model) -- (output);
    \end{tikzpicture}
    }
    \caption{The ML prediction model receives the box plots (for time and frequency domains) and machine configuration parameters to output the quality measure (surface roughness).}
    \label{fig:prediction_model}
\end{figure}
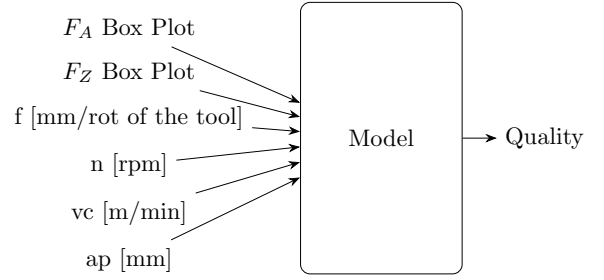

\subsection{ML Model Training and Evaluation}\label{sec:model_evaluations}
\begin{figure}[t]
    \centering
    \includegraphics[width=0.48\textwidth]{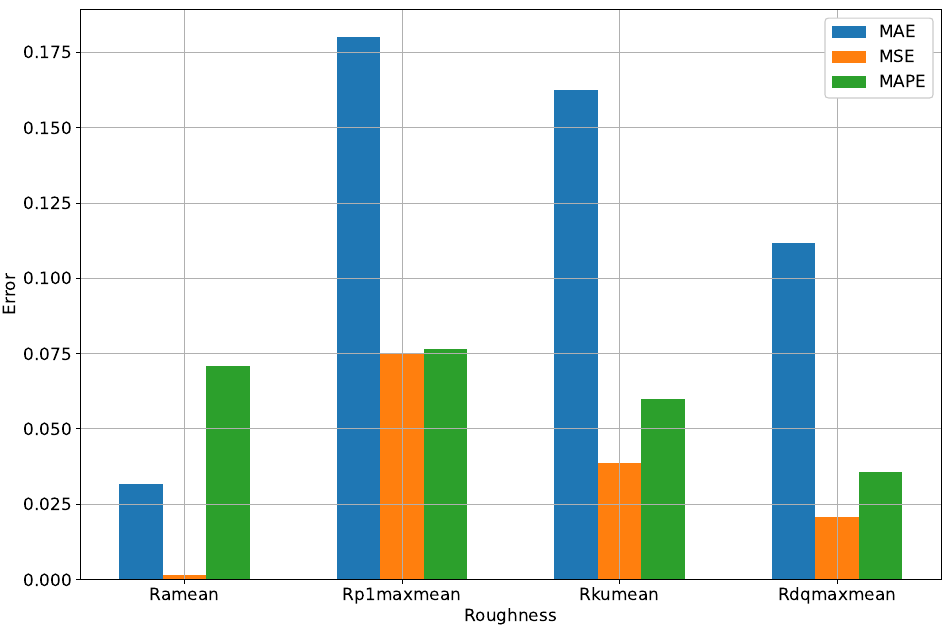}
    \caption{Predictive model test results with mean squared error (MSE), mean absolute error (MAE), and mean 
    absolute percentage error (MAPE). .}
    \label{fig:performance}
\end{figure}

We trained random forest regression models on the preprocessed dataset to predict various surface roughness measures.
The performance of the top four models is illustrated in Figure \ref{fig:performance}, evaluated using mean squared error (MSE)~\citep{DBLP:journals/jmlr/JinM23}, mean absolute error (MAE)~\citep{DBLP:journals/icl/SahaMSM22}, and mean absolute percentage error (MAPE)~\citep{DBLP:journals/ejivp/Maiseli19}.
MSE quantifies the average of squared prediction errors, making it sensitive to outliers, whereas MAE represents the average absolute errors.
MAPE, expressed as a percentage, measures the average absolute percent deviation from actual values, which is useful for comparing models across different scales.

Our models demonstrated proficiency in predicting \emph{Ramean}, \emph{Rp1maxmean}, \emph{Rkumean}, and \emph{Rdqmaxmean} surface roughness measures, achieving a MAPE under 8\% (refer to Figure~\ref{fig:performance}).
Notably, the model predicting \emph{Rdqmaxmean} reached a high-quality standard with a MAPE below 5\%.

\subsection{Assessing Gini Importance in Random Forest\\Regression Models to reduce number of sensors}\label{sec:gini_exp}
\begin{figure*}[t]
    \centering
    \includegraphics[width=0.85\textwidth]{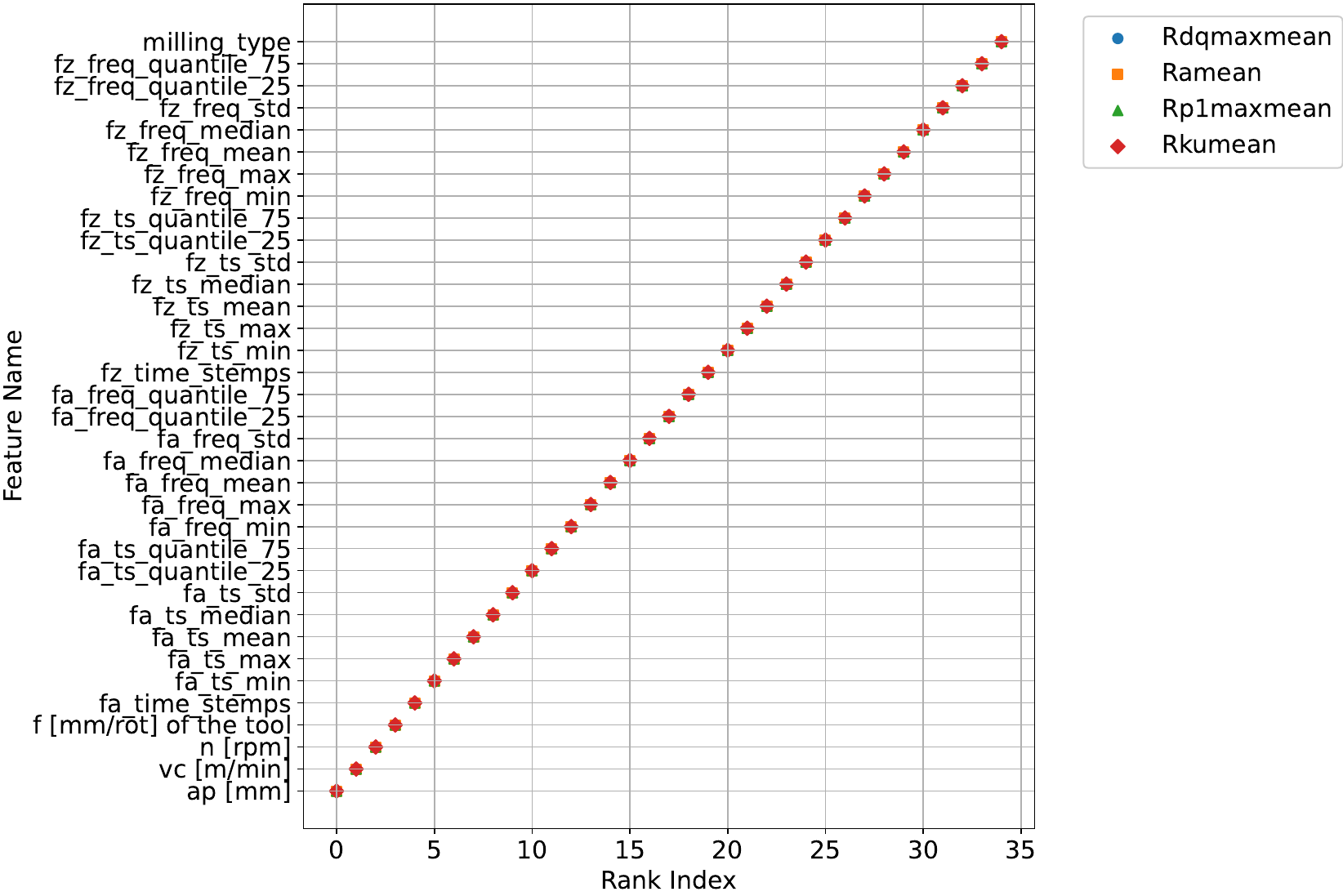}
    \caption{Feature Importance rankings across the different roughness prediction random forests (see labels). The feature ranking is identical for all prediction targets.}
    \label{fig:ginis}
\end{figure*}
Given the trained random forest regression models from Section~\ref{sec:model_evaluations}, we extract the gini importance for each feature (see Figure~\ref{fig:ginis}).
We observe that even though the models tried to predict different roughness types, the importance of the feature is the same in every model.
The most important features are the machine configuration parameters, followed by the box plot values of the active force ($F_a$), then the normal cutting force ($F_z$), and then the milling type (up milling/down milling).

Since all $F_a$- and $F_z$-related features are consistently ranked lower than the experiments' machine configurations, these sensors seem less relevant to the trained model.
We, therefore, remove all the $F_a$- and $F_z$-related features from our dataset and re-train the models.

The experimental results demonstrate that omitting the $F_a$ and $F_z$ sensors and concentrating solely on the machine configuration features leads to the development of more efficient predictive models.
Specifically, the performance improved for \emph{Ramean}, achieving a rate of $6.18\%$ compared to the previous $7.1\%$.
Similarly, the prediction accuracy for the surface roughness metric \emph{Rdqmaxmean} enhanced, reducing from $3.6\%$ to $3.1\%$.
Additionally, this refined dataset enables us to predict the \emph{Rzmean} surface roughness with a MAPE of $9.7\%$.
However, it is noteworthy that the performance metrics for both \emph{Rkumean} and \emph{Rp1maxmean} experienced a marginal decline of approximately $0.25\%$.

\subsection{Assessing the Significance of Feature Permutation Importance Across Various Data Subsets}
\begin{figure*}[t]
    \centering
    \includegraphics[width=0.85\textwidth]{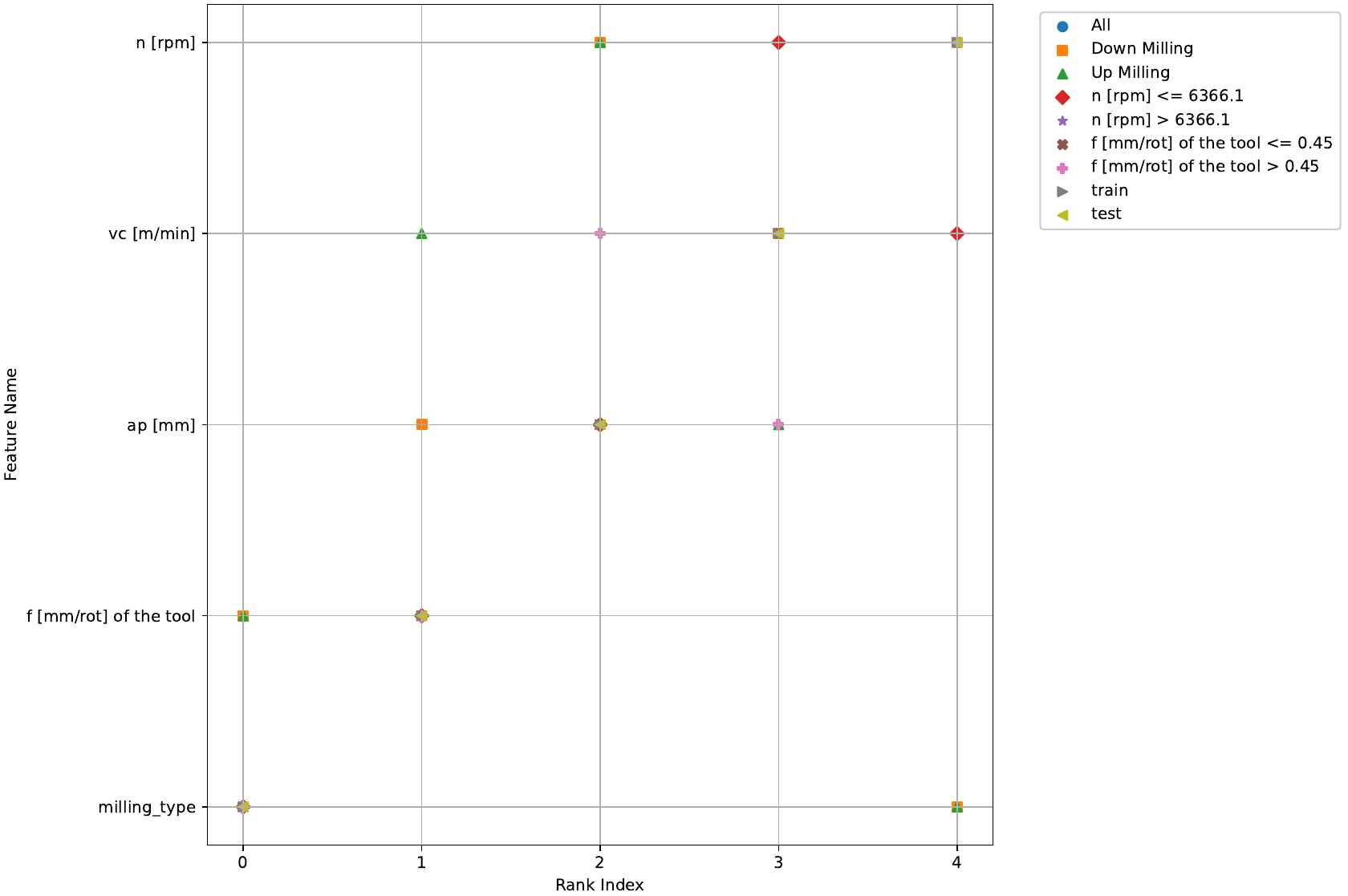}
    \centering
    \caption{Feature Permutation Importance across various data subsets.}
    \label{fig:fpi}
\end{figure*}
By analyzing the feature permutation importance of the trained random forest regression model for \emph{Rdqmaxmean} from Section~\ref{sec:gini_exp} across various data subsets, the experiment aims to understand how changes in individual features impact the model's predictions across different segments of the dataset.
A significant variation in feature permutation importance between the training and test sets could indicate that the models are overfitting.~\citep{DBLP:journals/kbs/BeltranRoyoLPR24}.

Figure~\ref{fig:fpi} suggests that the ML model exhibits relatively consistent behavior regarding how it values and responds to changes in each feature, regardless of the specific data subset being considered.
Note, the experiment is based on the premise that different explanation methods (gini importance in Figure~\ref{fig:ginis} vs. feature permutation importance in Figure~\ref{fig:fpi}) for ML models will result in varying interpretations, influenced by methodological differences, model complexity, and inherent biases in each approach~\citep{lozano2023comparison}.
We observe that there can be a small change in the feature rank depending on the data split investigated, e.g. the order of \textit{vc} and \textit{ap} is reversed for the two splits of \textit{f of the tool $\leq 0.45$}.
This shows that these features are relevant for the overall model and must be kept because they provide specific importance for the quality score prediction in certain subsets of the data.

\section{Discussion}\label{sec:discu}
Our case study demonstrates the advantages of using explainable ML techniques in the context of surface roughness quality prediction models within the manufacturing sector.
We can interpret the model predictions more effectively by utilizing explainability scores derived from feature importance analyses. This approach enables the identification and assessment of the significance of individual features in contributing to the overall predictive accuracy of the model.
For economic reasons~\citep{dornfeld2008sensors}, this allows us to remove sensors from the manufacturing machines while maintaining and even improving predictive accuracy for certain surface roughness values.
Additionally, utilizing these ML models as digital twins~\citep{DBLP:journals/bdcc/DAmicoAE23} for the corresponding physical machinery opens new avenues for employing parameter optimization methods.

Nevertheless, it is important to acknowledge the risk of overfitting, a potential issue amplified by the limited size of the dataset used~\citep{DBLP:journals/kbs/BeltranRoyoLPR24}.
This concern suggests that definitive conclusions about overfitting can be ascertained through future research employing, for instance, larger datasets.

\section{Conclusion}\label{sec:conclu}
This study applies explainable machine learning to predict milling surface roughness, utilizing random forest regression and feature permutation importance.
Key achievements include the creation of an accurate predictive model and the identification of non-essential sensors.
Eliminating these redundant sensors, especially those measuring normal cutting force, enhances the model's cost-effectiveness without sacrificing accuracy.
This work exemplifies how explainable ML can optimize manufacturing processes, combining improved performance with economic efficiency.

For future work, we plan to apply our approach to a range of materials and milling processes to assess the robustness and adaptability of our model.
This will include augmenting our dataset with synthetic data to improve ML training~\cite {DBLP:journals/corr/abs-2212-09317}.
Additionally, we intend to apply our predictive models to different machines of the same type, evaluating their performance in these new contexts.
Another particularly intriguing future work aspect of our research is leveraging these models within digital twins, facilitating advanced parameter optimization methods~\citep{soori2022review}.

\bibliography{ifacconf}
\end{document}